# Evaluating Actuators in a Purely Information-Theory Based Reward Model[*]

Wojciech Skaba
AGINAO
Trubadurow 11
80205 Gdansk, Poland
wojciech.skaba@aginao.com

*Abstract*—AGINAO builds its cognitive engine by applying self-programming techniques to create a hierarchy of interconnected codelets – the tiny pieces of code executed on a virtual machine. These basic processing units are evaluated for their applicability and fitness with a notion of reward calculated from self-information gain of binary partitioning of the codelet's input state-space. This approach, however, is useless for the evaluation of actuators. Instead, a model is proposed in which actuators are evaluated by measuring the impact that an activation of an effector, and consequently the feedback from the robot sensors, has on average reward received by the processing units.

## I. Introduction

Epigenetic robotics addresses the question of autonomous mental development by applying a control program – herein referred to as the *cognitive engine* – embedded in a physical robot that is interacting with the natural environment. The task may be characterized by the following features: (a) the engine doesn't have any a priori knowledge on the nature and meaning of its sensors and actuators, possibly not even on a distinction between being a sensor/actuator or not, (b) the computational resources of the engine are limited, the world partially observable, the flow of sensory data overwhelming, and the learning conducted in real time, (c) the objective of the task – referred to as the *global fitness function* – is a rather general in nature and not directly transferable to low level technical implementation; the fitness function is not given, either, thus must be carefully selected by the engine designer, (d) the engine must learn the rules of its embodiment and eventually the higher level mental skills, e.g., the understanding of a distinction between its body and the outer world, possibly the understanding of the existence of other independent intelligent agents. We would add yet another feature, not shared by all dealing with the domain: (e) the actuators and the sensors are accessible in raw format, i.e., without or with very little pre-/post-processing.

All of the listed above features, besides their obvious disadvantages, have one common advantage: we do not impose any restrictions on the structure of the emergent architecture, keeping in mind all the failures of the past approaches that have apparently assumed too optimistic constraints on the approach to artificial intelligence.

## II. The AGINAO Project

### A. The AGINAO Self-Programming Engine

A detailed presentation of the construction of the AGINAO self-programming engine is presented in [1]. What follows is an excerpt essential for the presentation of the paper thesis.

The AGINAO cognitive engine uses the NAO robot by Aldebaran Robotics as a testbed. The emergent architecture of the cognitive engine is constructed as a control program executed on a devised virtual machine (VM). Functionally, the control program is robot-embedded, though technically it is executed on a remote and more powerful host, connected to the robot via a wireless link; the robot to act as an interface to the outer world only. The architecture of the cognitive engine is open-ended and adjusted for real-time adaptability.

The operation of the cognitive engine is based on a conjecture that any algorithm, including the hypothesized algorithmic artificial mind, is computable on a Universal Turing-Machine (UTM) – the Church-Turing thesis, and that a suitably designed VM will have the flexibility and power comparable to a UTM. On the other hand, however, the target algorithm – any of the possibly infinitely many accomplishable implementations – is unknown.

This approach assumes a somewhat random construction of the target algorithm, and its evaluation during the learning process. The creation of such an algorithm as a single piece of code is theoretically sound but at best intractable [2]. Various methods have been attempted to deal with the incomputability-intractability question, to list a few [3], [4]. AGINAO builds its cognitive engine as a hierarchy of interconnected data-structures, named *concepts*, each with a built-in piece of executable code, named *codelet*. The unidirectional links between concepts specify both the order the concepts are applied (executed) and the data-flow between the concepts. Typically, the output of each concept becomes an input of many other of its descendents.

The cognitive engine is not a neural network, however. Concepts from the hierarchy stand for a repository of programs to be executed on the mentioned above VM. The system is capable of running many threads concurrently, not excluding a concurrent execution of multiple copies of the same concept's codelet, most likely processing different data.

There are also special-type predefined atomic concepts connected to the robot sensors and actuators. For the cognitive engine, they look much like *regular* concepts but their implementation lacks a codelet; a hidden [for the cognitive engine] functionality is encoded instead. The *sensory* concepts are integrated to the hierarchy as the roots, while the *actuator* concepts as the terminal leaves.

During the continuous operation of the cognitive engine, the candidates for regular concepts are generated by a random process. The created codelets are tiny programs, consisting of typically 4–10 instructions (symbols of the input alphabet) of the VM. The instruction-set resembles those of early 1980s microprocessors. By virtue of their compactness, the codelets are easily tractable. Starting from its creation, a codelet passes through the following mutistep process of evaluating its validity and applicability, resulting in the majority of the preliminary codelets to be discarded:

- *Heuristic-search in program-space* applies tricky heuristics to sort-out pieces of code obviously useless and imposes many straightforward constraints. This step is enforced before a codelet is integrated into the hierarchy.

- Runtime-error detection, performed during codelet execution, catches fatal errors and protects against running out of computational resources, e.g., infinitely looping. A common type of a fatal error is an attempt to read/write data out of scope.

- Evaluation of the concepts for their fitness to the overall structure of the hierarchy discards concepts that are rarely used and those having low value.

The last condition entails the necessity of implementing a measurement of concept's value. AGINAO applies a sort of temporal-difference learning (TD-learning) where concept's value is computed from both the immediate-reward and the discounted future-reward. The expected depth of the concept hierarchy, however, suppresses and consequently invalidates the discounted reward on long distances, a phenomenon known as convergence to a suboptimal policy [5].

On the other hand, the discounted reward may be beneficial on short distances. This may be compared to driving a car. The driver may estimate the optimal steering policy for a few time-steps ahead, by observing the obstacles within the visible distance. With every time step, the horizon also moves one step forward, and a new policy may be obtained. On the other hand, anything that happens closer to the destination of the journey has no effect on the evaluation of the current policy, even if the opposite traffic has been influenced by that, i.e., carries some information. It may be said that solving the main problem (getting to the destination) is biased by a carefully selected immediate-reward function. AGINAO introduces a notion of intrinsic-reward (as immediate-reward) based exclusively on information-theory and calculated from self-information gain of binary partitioning of the codelet's input state-space.

### B. Binary Space-Partitioning and Computation of the Reward

The application of binary space-partitioning for the estimation of the immediate-reward was first introduced in [6]. Fig. 1 shows the idea of binary space-partitioning.

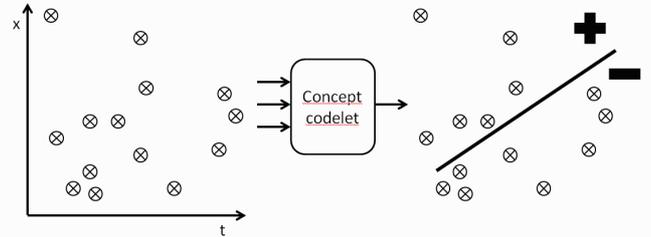

Figure 1. Binary space-partitioning

AGINAO uses a unified format for data interchange: a variable-size vector of integers. The input of a concept may be depicted as discrete-time points of a spatial-temporal state-space. The task of the codelet is to separate the positive (pattern-matching) examples from the negative (non-pattern-matching). The definition covers temporal patterns, as well.

An illustrative example might be to consider a task of detecting the letter **T**. Imagine that the inputs of a 2-input concept are interpreted as: (1) to signal a detection of a horizontal bar; (2) to signal a detection of a vertical bar,. The inputs are connected to the outputs of some lower level concepts. If non of the bars were detected in the visual scene, the inputs will not receive data. The values of the input vectors represent the coordinates of the detected bars. The task of the codelet is to take the coordinates and check if the spatial arrangement of the bars matches the shape of the letter **T**. Moreover, the bars must coexist in real time. If a match is found, the output will forward some data vector (possibly letter coordinates) to other concepts. Otherwise, the execution of that thread would be abandoned. Possibly, another concept applied to the same state space could detect letter **L**.

From the proportion of the $N_{pos}$ positive and $N_{neg}$ negative examples, the probability (variable in time) of encountering a positive example is computed:

$$p = N_{pos}/(N_{pos}+N_{neg}) \tag{1}$$

The *self-information* [7], i.e., the amount of information (in bits) provided by an event of getting a positive example, may be extracted:

$$I = -log_2(p) \tag{2}$$

Since the probability of getting a positive example is $p$, we get a measure of mean reward, as average information-gain:

$$r = -p\, log_2(p) \tag{3}$$

The notion of self-information expresses our intuitive feeling that the more rare an event is, the more information it entails. On the other hand, expecting a reward from a rare event is rather risky. The reward function maximizes at p=1/e (Fig. 2). The mean reward, rather than the reward provided by a positive example, is interpreted as immediate-reward for TD-learning, for we want to estimate the value of a concept.

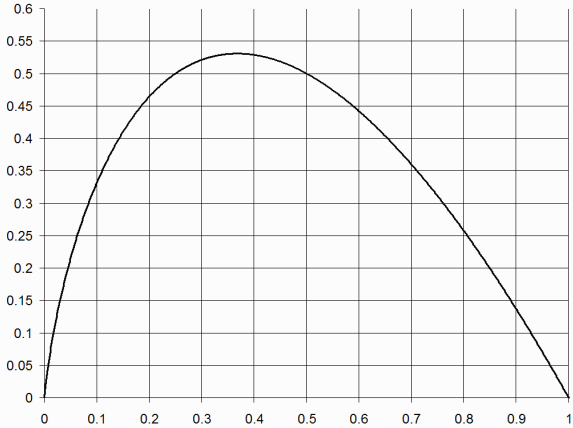

Figure 2.  $-p \log_2(p)$

Self-information is also a special case of Kullback-Leibler distance from a Kronecker delta representing the matching pattern to the probability distribution [6], [8].

Worth noting are the following observations:

- Since the partitioning is performed by a codelet, and the power of the VM is equivalent to a UTM, a concept is capable of detecting virtually any regularity in the input data.

- While the algorithm of data partitioning is codelet specific, the resulting probabilities, and hence the computed rewards, are input-data specific. Consequently, for different state-spaces we would get different measures of reward for a given codelet.

- A codelet may be regarded as a binary non-linear and unsupervised classifier that is supposed to carry-out not *the best* but *any* partitioning. As a rule, for each state-space many different codelets will be applied concurrently. In a continuous process of adding, evaluating and removing concepts, the cognitive engine attempts to maintain a subset of the most valuable concepts for each state-space.

- Only a detection of a positive example is rewarded. A failure to match a pattern simply cancels the thread and skips the TD-learning update step.

C. *Global Fitness Function*

AGINAO employs a notion of [a global] *average reward per time step*. Instead, however, of using the discrete time-step of a Markov Decision Process (MDP), the following function, based on real-time events and exponential decay, is applied:

$$R_t = r_t + R_{t_0} e^{-\rho(t-t_0)} \quad (4)$$

where $R_t$ is the computed average reward at current real-time $t$, $R_{t_0}$ is the average reward computed at time $t_0$ in the past (the last time it was computed), $r_t$ is the immediate-reward at the current time $t$, $\rho$ is a positive constant to control the rate of decay. The average reward is computed as a single value shared by all processes of the cognitive engine and updated every time the immediate reward is received.

The global fitness function is defined as the maximization of the average reward per time step.

D. *Artificial Economics*

The cognitive engine as a whole behaves like a *complex adaptive system*. The concepts operate as interacting adaptive agents, collaborating and/or competing, and fighting for the computational resources, according to the rules of implemented *artificial economics*. At every time there is an overabundance of the codelets requesting access to the VM and awaiting in a priority queue. Many of these codelets, possibly most of them, will never get serviced, exactly like in nature, most individuals will never breed offspring.

For the topic of this paper, it is sufficient to mention that each codelet's thread is assigned: (a) *priority* – a positive integer governing the order of execution of the codelets, (b) *expiration time* – an unconditional deadline for every thread, (c) *resources* – a positive integer limit of the utilization of the VM processing time (in steps). The latter also solves the challenge of dealing with the halting problem.

The resources are expresses in the same units as the immediate-reward, subject to a normalization coefficient $\beta>0$. A pattern-matching thread is rewarded with extra resources $s$ computed as $s=\beta I$, where $I$ is the self-information gain in bits. Depending on the amount of the available resources, the thread will pass execution and its output data to the descendents, or abandon.

III. EVALUATING ACTUATORS

A. *Problem Specification*

From the cognitive-engine perspective, an actuator is perceived as a concept with known number of inputs and known minimum sizes of each input (minimum number of integers of input vector). The meanings of the individual actuators and the meanings of their inputs are unknown and are supposed to be discovered during the learning process.

To be executed and evaluated, an actuator-concept must be integrated into the concept hierarchy. The integration means connecting the input(s) of an actuator-concept to the output(s) of the regular ones. Each individual actuator is represented by a single atomic actuator-concept template. An actuator, however, may be potentially linked concurrently to different location of the concept hierarchy, to mean that it could be executed in different contexts. On the other hand, it wouldn't be very beneficial to evaluate a given actuator with a single-value parameter, shared by all different contexts. To solve this problem, each atomic actuator-concept *T* is copied before being connected to the hierarchy; then the copy $T_i$ is evaluated independently in each context. A future possible removal of a one of the copies, due to low value, will not invalidate the other copies. This approach, however, implies the question of resolving some conflicts, discussed below.

*B. TD-learning Rule for Actuator*

For regular concepts, the value $Q_{i,t}$ of action $a_i$ at time *t* is updated according to the following TD-learning rule:

$$Q_{i,t} = Q_{i,t} + \alpha \left[ r_{i,t+1} + \gamma \overline{V}_{i,t+1} - Q_{i,t} \right] \quad (5)$$

where $r_{i,t+1}$ is the immediate reward at time *t+1*, *α* is the learning rate, *γ* is the discount factor, and *V* is the weighted average of values of all actions of the concept the action $a_i$ points to, computed from the equation:

$$\overline{V} = \frac{\sum_j p_j Q_j}{\sum_j p_j} \quad (6)$$

Where $p_i$ is defined by (1). The probability of selecting action $a_i$ at time *t* is defined as:

$$P(a_{i,t}) = \frac{Q_{i,t}}{\sum_j Q_{j,t}} \quad (7)$$

The above definition of TD-learning rule is not applicable for the actuators for many reasons: (a) actuator-concepts do not contain a codelet that computes the immediate-reward, (b) the actuator-concepts do not really partition the state-space, but merely function as a physical-actuator proxy, (c) the computation of the weighted average value *V* is impossible due to the fact, that actuator-concepts are terminal-leaves and do not connect to next actions/descendents. What follows, the immediate-reward must be removed from the TD-learning rule equation, and the weighted average value must be replaced with a value of an actuator-concept evaluated independently. Effectively, for the next-to-last [the terminal] concept in a chain we get the following formula:

$$Q_{i,t} = Q_{i,t} + \alpha \left[ \gamma A_{i,t+1} - Q_{i,t} \right] \quad (8)$$

where $A_{i,t+1}$ is the independently evaluated estimation of actuator-concept at time-step *t+1*.

*C. Basic Idea of Actuator-Concept Evaluation*

The idea of evaluating the actuator-concepts is based on an assumption that, rather than introducing a new source of reward, we would exploit the already defined information-theory based reward.

Let us consider first what happens when an actuator is activated, and take a robot's arm movement as example. We anticipate that the resulting impact on the environment will be reflected in robot's sensors. To focus attention, consider the visual sensory system. The robot's arm repositioning may cross the visual field, or not. It will result in observing/detecting a pattern in the former case, and not in the latter case. According to the discussed above rules of artificial economics, only the former case will be rewarded with information-gain and resources. What follows, we want to maintain the concept structures responsible for the observed feedback and discard those which effect on the sensors is unknown.

*D. Actuator's Cost Function*

Before continuing, we have to make yet another remark. Every thread is given a resources limit. For regular concepts, this is interpreted as the limit for the VM processing time. As for the actuators, since we want to observe the rules of artificial economics, we have to convert the resources to the physical energy-consumption equivalents of the effectors, herein referred to as the *cost*. The mind-body energetic correspondence – natural in the living organisms – must be simulated in case of the robotic embodiment.

The cost of activating a given actuator is not fixed, but depends on its input parameters, like range of the arm movement. We don't know the meanings of the actuator-concept inputs, however, not even if increasing the input's vector value increases the movement range and the cost of actuator's activation. The details of internal implementation are hidden for the cognitive engine.

Yet another question arises: a typical cost of actuator activation – if expressed in the same units as the resources – is much beyond the average resources assignment of a thread. This observation seems quite obvious, if one compares energy dissipation of a robot motor and that of executing a few hundred instructions by a contemporary PC. What follows, an activation of an actuator must result from many consecutive requests to the actuator-concepts of a given actuator.

Let's assume that the cost of activating an actuator $T_i$ is given by function $C_i(x_1,...,x_n)$, where $x_1,...,x_n$ are the input parameters (vectors of integers, the internal data format), $n>0$; With each request to execute actuator $T_i$ some resources $s_{i,j}$ are passed to it. A naïve approach to deal with the activation problem would be based on implementing an actuator as a resources integrator. Once the combined resources have exceeded the cost threshold, the actuator would be activated:

$$\sum_j s_{i,j} > C_i(x_1,...,x_n) \quad (9)$$

We encounter a problem here, however, if the consecutive requests differ in the input values $(x_1,...,x_n)$, i.e., if the actuator receives contradicting requests, like: move the arm left and right. Any summing of the resources in such case is senseless, as the requests come most likely from different contexts.

Attempts to overcome this problem by, e.g., computing the weighted average of the request-commands' inputs, or computing independent sums for each set of input parameters, until any of them exceeds the threshold, seems inappropriate. If, on the other hand, the input parameters differ insignificantly, we have to figure out whether count them together or separately. If the summing takes too long, the very first requests seem somewhat depreciated, and should not be taken into account.

We propose a solution based on a notion of probability of executing an actuator, computed as:

$$p_{s_{i,j}} = \frac{s_{i,j}}{C_i(x_1,...,x_n)} \quad (10)$$

This approach solves all the problems mentioned above. It also assigns higher probability to stronger signals (resources), that might differ significantly if the requests come from different contexts. A higher probability is also assigned to lower cost actions, e.g., an arm movement on a shorter distance is more likely.

One might observe that if the consecutive input vectors $(x_1,...,x_n)$[1] are fixed [constant], if $S = \{s_{i,1},...,s_{i,m}\}$, $m>0$, and:

$$\sum_S s_{i,j} = C_i(x_1,...,x_n) \quad (11)$$

then:

$$\sum_S p_{s_{i,j}} = \frac{1}{C_i(x_1,...,x_n)} \sum_S s_{i,j} = 1 \quad (12)$$

that means that on average we get the same frequency of activating an actuator as in the case of a resources integrator (9), i.e., exactly what would be expected.

*E. Evaluating the Actuator Concept*

Now imagine that an actuator was activated and we expect that – after some delay, currently unknown, but counted in tens or hundreds of milliseconds rather than nanoseconds – we receive a feedback, that will be reflected in robot's sensory system as increase of immediate-reward. At this point, we have completely no idea in what portion of the concept hierarchy would the impact be observed. The only available measure of the impact is the change of average reward per time step, defined by (4).

Let $X_{i,to} = C_i(x_1,...,x_n)/\beta$ at time $t_0$, i.e., the time[2] the actuator $T_i$ was activated. $X_{i,to}$ is just the cost recorded at time $t_0$ and expressed in bits, i.e., the cost related to the current input vector

---

[1] It is a vector of vectors of integers.
[2] Again, we are using alternately the notion of time $t$ as time-step of MDP, or as real-time. The distinction should be read from the context.

$(x_1,...,x_n)$. At time $t>t_0$, i.e., after $t-t_0$ delay, the following value update rule is performed:

$$A_{i,t} = A_{i,t} + \alpha[\delta(R_t - R_{t_0}) - X_{i,t_0}] \quad (13)$$

where $R_t$ are $R_{t0}$ are average rewards per time step computed at times $t$ and $t_0$, respectively, $\alpha$ is the learning rate, $1 \geq \delta > 0$ is a normalization coefficient. The value $R_t - R_{t0}$ may be negative. Even if $R_t - R_{t0} = 0$, the $A_{i,t}$ will decrease, due to the cost term. If $A_{i,t}$ goes below a predefined positive threshold limit, the related actuator-concept will be removed.

The $\delta$ coefficient was introduced to take into account the fact, that the time delay $t-t_0$ is relatively long. What follows, many concurrently activated actuators might have influenced the change in average reward, before the update rule (13) was performed. We have to share the change of average reward among all active actuator-concepts. Consequently, the $\delta$ coefficient must be implemented as a function $\delta(t)$ rather than a constant. In the current implementation, $\delta(t)$ is defined as $1/N$, where $N>0$ is the number of actuator-concept that have activated the actuators but have not passed the value-update step yet.

*F. Possible Future Improvments of the Evaluation Method*

The presented above method of actuator-concept evaluation is quite limited for the following reasons:

- even if no actuator is activated, the average reward is a rather volatile function of time, due to both the internal cognitive processes and the events in the environment not related to robot actuators; what follows, not all changes in the average reward level may be attributed to actuator's activation,

- the temporal delay $t-t_0$ is unknown and most likely variable. The delay is caused by at least the following three phenomena: (a) the inertia of the actuator, (b) the time the environment reacts to an action, (c) the time the sensory data propagates through the concept hierarchy. The latter to mean that the average-reward change function is closer to a normal or Boltzmann distribution rather than a single peak, for the extra reward may come from many levels of the concept hierarchy. An attempt to determine experimentally a typical delay is presented in section on experimentation (Fig. 3).

We have to highlight that this study focuses on simple reflexes, expected to happen within a couple of seconds at most. We do not consider feedbacks that would require sophisticated world-modeling. The purpose of the learning is to establish the functionality of the actuators, and then *suppress* further evaluation. Once the functionality is established, the value of each actuator-concept $A_i$ is propagated as discounted reward via (8), and then via (5).

The effect of suppression is currently implemented by analogy to the method of estimating the probability of exploration[3], presented in [1]. Let's assume that for an actuator $T$ there exist copies $T_1,...,T_n$, $n≥0$, with the current value estimates $A_1,...,A_n$. The probability of exploration step (creating a new copy $T_{n+1}$) is computed as:

$$P(Exploration) \propto \frac{A_{const}}{A_{const} + \sum_i A_i} \quad (14)$$

where $A_{const}$ is a predefined constant (estimated experimentally). With the increasing number $n$, and increasing value of each $A_i$, the probability of adding a new actuator-concept's copy $T_i$ decreases, and effectively, the evaluation process for a given actuator $T$ becomes suppressed. Experiments have been conducted with both limited maximum value of $n$ (currently 50), and unlimited. In the former case, the newly created actuator-concept's copy replaces the one with the lowest current evaluation $A_i$.

An improved model of actuator-concept evaluation, that hasn't been implemented yet, would focus on an attempt to extract a subset of concepts actually influenced by each actuator-concept $T_i$ independently, and determine the temporal delay, then apply the (13) update rule based on the selected subset, now with $δ=1$. This approach would involve the following steps:

- selection of a set of {concept, time-delay} hypotheses, independently for each actuator-concept $T_i$, most likely with many hypotheses per concept differing in time delay,
- learning the Bayes net and eliminating the false hypotheses,
- applying the reward changes of only the extracted subset of concept for the update rule, now without the $δ$ coefficient.

The task, as presented above, is too computationally intensive to be performed on all concepts for all actuators in real time. An approach must be found to make the problem more tractable.

## IV. EXPERIMENTS

Fig. 3 presents the results of finding experimentally the impact that a change in robot's visual field has on average-reward-per-time-step function, especially an attempt to determine a typical delay of a maximum response. The experiment was conducted by running the real robot on the real visual data and concurrently overlapping a disturbance simulated in software. The data collection step was preceded by initial undisturbed learning (50 sec). Then, without interrupting the learning, the visual field was stimulated with strong signal lasting 100 ms, followed by a 2-sec break, repeated 30 times. The average reward level was collected with 10 ms resolution and summed over the repeated stimulation periods. The maximum of the response seems to be at around 300 ms from the beginning of the visual sensory stimulation. The experimentally established delay may be used as an a priori probability of the maximum response delay for the Bayesian learning.

---

[3] According to a convention assumed in [1] and [6], we would use the name *exploration* for adding a new action, and *exploitation* for selecting any available action, not only the most greedy one.

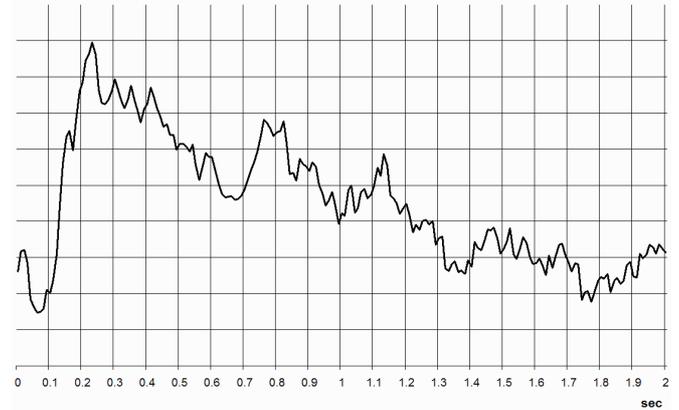

Figure 3. Average-reward change

## V. SUMMARY

This paper presented a method of evaluating the actuators as an extension of the earlier developed method of evaluating the regular concepts processing sensory data. Both approaches exploit a purely information-theory based notion of the self-information gain computed from binary partitioning of concept's input state-space. Without introducing a new source of reward, and adopting the paradigm of maximizing the average reward per time step, the actuators are assessed according to the impact they have on the global reward.